\pdfoutput=1

\documentclass[11pt]{article}

\usepackage{acl}
\usepackage{graphicx}
\usepackage{bbm}
\usepackage{subfigure}
\usepackage{stfloats}

\usepackage{hyperref}

\usepackage{times}
\usepackage{latexsym}
\usepackage{multirow}
\usepackage[T1]{fontenc}

\usepackage[utf8]{inputenc}

\usepackage{microtype}

\usepackage{inconsolata}

\usepackage{ulem} 

\title{HKUST at SemEval-2023 Task 1: Visual Word Sense
Disambiguation with Context Augmentation and Visual Assistance}

\author{Zhuohao Yin, Xin Huang \\
  Hong Kong University of Science and Technology \\ 
  \texttt{\{zyinad, xhuangcx\}@connect.ust.hk} }

\begin{document}
\maketitle

\begin{abstract}
Visual Word Sense Disambiguation (VWSD) is a multi-modal task that aims to select, among a batch of candidate images, the one that best entails the target word's meaning within a limited context\footnote[1]{This work is not among the participants of SemEval-2023 Task 1.}. In this paper, we propose a multi-modal retrieval framework that maximally leverages pretrained Vision-Language models, as well as open knowledge bases and datasets. Our system consists of the following key components: (1) Gloss matching: a pretrained bi-encoder model \cite{blevins-zettlemoyer-2020-moving} is used to match contexts with proper senses of the target words; (2) Prompting: matched glosses and other textual information, such as synonyms, are incorporated using a prompting template; (3) Image retrieval: semantically matching images are retrieved from large open datasets using prompts as queries; (4) Modality fusion: contextual information from different modalities are fused and used for prediction. Although our system does not produce the most competitive results at SemEval-2023 Task 1 \cite{raganato-etal-2023-semeval}, we are still able to beat nearly half of the teams. More importantly, our experiments reveal acute insights for the field of Word Sense Disambiguation (WSD) and multi-modal learning. Our code is available on \href{https://github.com/Thomas-YIN/SemEval-2023-Task1}{GitHub}.

\end{abstract}

\section{Introduction}

Polysemy is a fascinating yet tricky characteristic of human languages. Word Sense Disambiguation (WSD) aims to empower AI systems to correctly discriminate senses of polysemic words under different contexts. Under traditional WSD settings, the problem is a classification task taking a context as input and all available senses of the target word as the labels. The accurate disambiguation of complex word senses is crucial for many downstream applications, such as information retrieval in search engines, machine translation, etc. 

Visual Word Sense Disambiguation (VWSD) can be seen as an extended multi-modal task over traditional WSD. It was first formally proposed by \cite{gella-etal-2016-unsupervised}, where the task is to match images depicting an action to the correct sense of the associated polysemic verb. Under the SemEval-2023 Task-1 setting \cite{raganato-etal-2023-semeval}, the task is different in that we are asked to return the image associated with the correct sense given a context, a target word, and a set of 10 candidate images. At a macro level, this is a image-text matching problem which is studied in a range of multi-modal learning frameworks \cite{radford2021learning, kim2021vilt, li2022blip}. However, VWSD is evidently a more difficult task as the dataset was formed by deliberately putting images depicting different senses of the same word together as candidates \cite{raganato-etal-2023-semeval}. This poses a much more challenging task that requires the model to identify the subtle differences between senses, which are semantically much closer to each other, as opposed to discriminating among a set of largely heterogeneous candidates. Another challenge lies in the limited context length provided. For each target word, the context consists of two to three words, including the target word itself. 

Thus, the VWSD task can be solved using a disambiguate-and-discriminate strategy. In the disambiguation stage, we embed the given context and available senses of the target word in an external knowledge base and select the most similar one measured by cosine similarity. We are then able to form an augmented context by fitting the matched sense into a prompting template. The augmented context gains rich semantic information and is suitable to serve as the input for image-text matching. Additionally, we resort to visual assistance by using the augmented context to retrieve images from open datasets. This approach mimics the way humans learn to recognize rare objects or sophisticated concepts. In the discrimination stage, we adopt a pretrained CLIP \cite{radford2021learning} to encode the augmented context, the retrieved images, and the candidate images. To leverage multi-modal information, we further incorporate a modality fuser to integrate information from the augmented context and the retrieved images to obtain a fused feature representation. Eventually, classifications are made using a 1-nearest-neighbor approach by measuring the cosine similarity between the fused feature and the candidate image features.

\section{Related Work}

Traditional WSD methods can be roughly divided into two categories, i.e., knowledge-based approach and supervised approach. A representative work of knowledge-based approach is the Lesk algorithm \cite{10.1145/318723.318728}, whose idea is to determine a word's sense by evaluating the overlap between its context and its signature. A signature of a sense is the combination of the gloss and the example sentences. Such a simple algorithm performs decently well with only a sense inventory, such as the WordNet \cite{wordnet} and the BabelNet \cite{NAVIGLI2012217}
. In more recent years, researchers have developed supervised learning algorithms on the task of WSD. Some works formulate it as a classification problem where the objective is $\mathcal{L}(w, c, s)$, where $w$, $c$, $s$ are respectively the target word, the context, and the sense. Such models usually leverage a pretrained language model to obtain context embeddings and feed them to a neural network or Transformer to produce a probability distribution over available sense \cite{bevilacqua-navigli-2019-quasi, hadiwinoto-etal-2019-improved}. Another line of works embed both the contexts and the senses into the latent space and perform 1-nearest-neighbor classification in terms of cosine similarity \cite{blevins-zettlemoyer-2020-moving}. In this way, we are able to enrich the senses by retrieving supplementary information from various knowledge bases like Wikipedia, instead of merely treating them as classification labels \cite{Scarlini_Pasini_Navigli_2020}.

CLIP \cite{radford2021learning} stands for Contrastive Language-Image Pre-Training is a Vision-Language model that embeds texts and images into the same latent space. It is trained to accomplish the objective that matched image-text pairs should have close representations in the latent space. Thus, it is inherently suitable for performing image-text matching tasks in the same way \citealt{blevins-zettlemoyer-2020-moving} match senses and contexts. The organizers of SemEval-2023 Task 1 used a pretrained CLIP to produce baseline results by embedding the raw contexts and candidate images. However, the VWSD task differs from regular text-image matching task in two major aspects. First, the text input typically consists of 2 to 3 words, where there is 1 target word with the rest serving as the context while CLIP's training data generally contains longer sentences describing the images. Second, VWSD target words tend to be rarely-used words such as latin names of plants and animals, which can be out of distribution for CLIP. Consequently, using CLIP without additional knowledge may not yield optimal results.

\section{Method}

\begin{figure*}
    \centering
    \includegraphics[width=\textwidth]{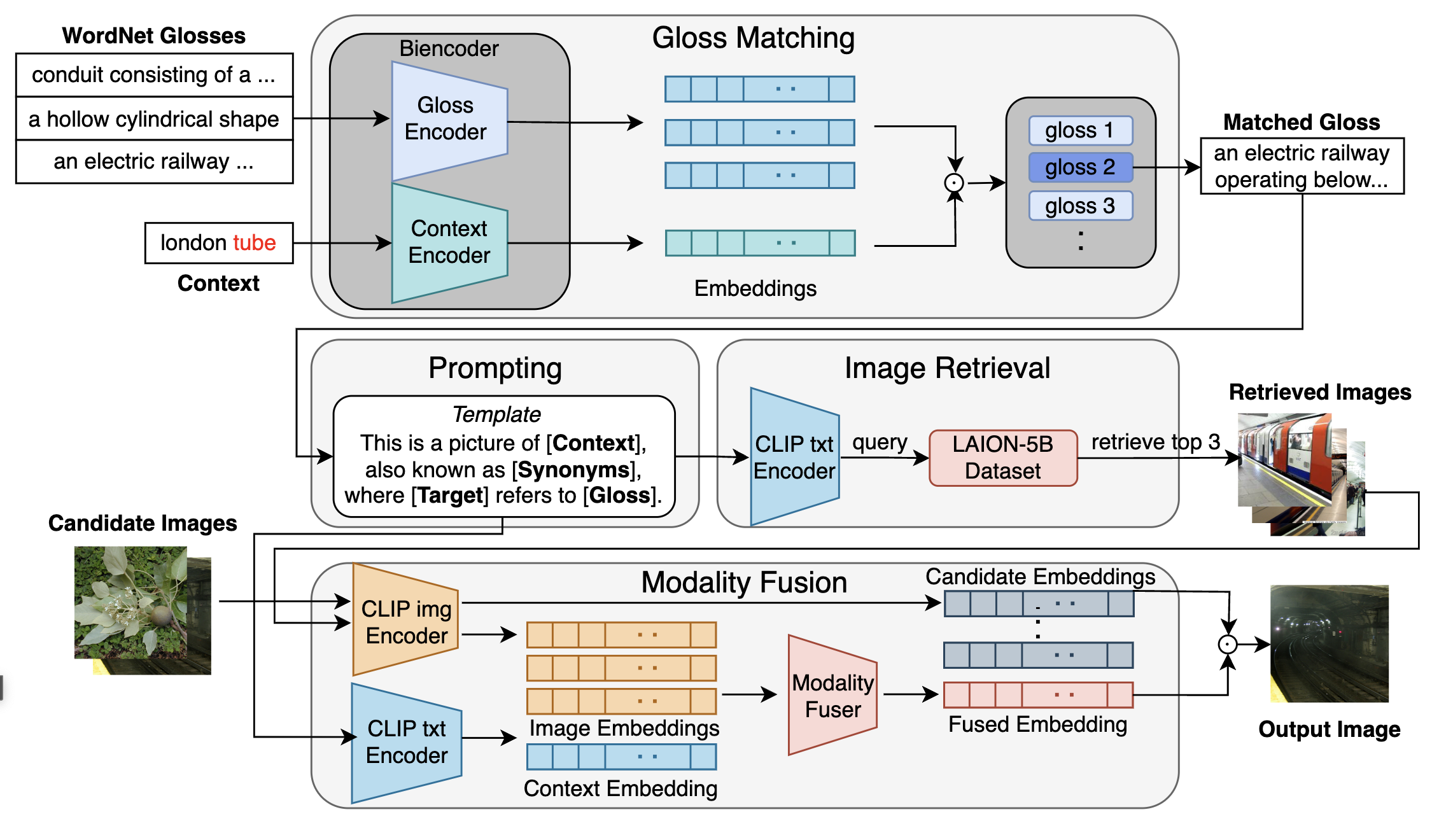}
    \caption{An overview of our system.}
    \label{fig:overview}
\end{figure*}

Our system is primarily composed of four modules as illustrated in Figure \ref{fig:overview}: (1) Gloss Matching; (2) Prompting; (3) Image Retrieval; (4) Modality Fusion. 

\subsection{Gloss Matching}

As an augmentation of the limited contexts provided, we first employ the Bi-Encoder Model by \citet{blevins-zettlemoyer-2020-moving} to find matching glosses for the target words from an external sense inventory. The BEM maps both the contexts and available glosses in the sense inventory, and returns the gloss whose embedding is the nearest to the context measured by cosine similarity. We adopt WordNet \cite{wordnet} as our sense inventory. 

\subsection{Context Augmentation through Prompting}

Once we complete gloss matching, a WordNet synset is matched for each example. We use the synset definition as the gloss and other words in the synset as synonyms. Subsequently, we incorporate the heterogeneous information using the following prompting template. An example is appended below it.

\vspace{5pt}

\fbox{
\begin{minipage}{18em}
\texttt{This is a picture of \{Context\}, also known as \{Synonyms\}, where \{Target word\} refers to \{Definition\}.}
\end{minipage}}
\vspace{5pt}

\fbox{
\begin{minipage}{18em}
\texttt{This is a picture of biro pen, also known as ballpoint, ballpoint\_pen, ballpen, Biro, where biro refers to a pen that has a small metal ball as the point of transfer of ink to paper.}
\end{minipage}}
\vspace{5pt}

For words that are too rare to be documented in WordNet (i.e. no matching synset), we simply leave out the latter part of the prompt and only keep \texttt{"This is a picture of"}. This practice follows \citet{radford2021learning}'s zero-shot experiments, which has been reported to bridge the distribution gap as the training texts of CLIP are rarely short phrases containing two to three words.

\subsection{Image Retrieval}

We also utilize external visual information using CLIP Retrieval \cite{beaumont-2022-clip-retrieval}, a toolkit that retrieves matching images from the LAION-5B dataset \cite{schuhmann2022laionb} given a text input. We sample a batch of top 3 images per context by inputting our structured prompt to CLIP Retrieval as the query. In total, 38607 images for the training set and 1389 images for the testing set are retrieved.

\subsection{Modality Fusion}

The raw context is enriched with both textual and visual information using external resources. To leverage the multi-modal information for classification, we design several modality fusers to merge the rich contextual information.

\textbf{Average fuser} has no trainable parameters. The Average fuser is a form of model ensemble. It encodes the augmented context and retrieved images with the CLIP text and image encoders to produce separate embeddings. We take the dot product of each embedding and the 10 candidate embeddings to compute cosine similarities, and apply softmax to produce 4 sets of probabilities. We then take the average of the probabilities to obtain the final classification result.

\textbf{MLP fuser} employs a multi-layer perceptron for modality interaction. The augmented context and retrieved images are again encoded first with CLIP. Next, the 4 embedding vectors of length 512 are concatenated on dimension 1 to form a vector of length 2048. The MLP takes as input the concatenated embedding and outputs a fused embedding of length 512. The final classification steps are the same.

\textbf{Transformer fuser} employs 2 layers of Transformer encoders, each with 8 attention heads. Again, we first encode the augmented context and retrieved images using CLIP. Next, the 4 embeddings are stacked to form a tensor of shape (4, 512), which is fed into the Transformer encoders. We then sum the outputs along dimension 0 to obtain the fused embedding. The classification steps stay the same.

\section{Experimental Setup}

Although SemEval-2023 Task 1 was published as a multilingual task which included Italian and Farsi in the test set, we only deal with English examples as examples in foreign languages can be first translated in English and processed in the same way.
We excluded several samples whose context contains special characters that cause IO issues, specifically 43 samples from training data and 3 samples from test data. We hold out 869 samples from the training set to form a validation set. 

The only trainable component within our system is the modality fuser while we use pretrained BEM \cite{blevins-zettlemoyer-2020-moving} and CLIP \cite{radford2021learning} both in zero-shot settings. The MLP fuser is trained for 3 epochs with learning rate 5e-5. The Transformer fuser is trained for 5 epochs with learning rate 3e-6. We also attempted to add a learnable linear projection layer for each intermediate embedding. However, the results are suboptimal and thus are discarded.

\section{Results and Discussion}

The evaluation metrics we use include hit rate at 1 and mean reciprocal rank (MRR) as suggested by \citet{raganato-etal-2023-semeval}. Formally, denote $r = [r_1, r_2, ..., r_n]$ to be the predictions for the total $n$ samples, where $r_i$ is the output ranking of the gold image label of the i-th sample. Then hit rate is defined as:

\begin{equation}
\mathrm{HIT@1} = {1 \over n} \sum_{i=1}^n \mathbbm{1}(r_i) ,
\end{equation}

where $\mathbbm{1}(r_i) = 1 $ if $r_i=1$, i.e., the gold image is ranked the first, otherwise $0$. MRR is defined as:

\begin{equation}
\mathrm{MRR} = {1 \over n} \sum_{i=1}^n {1 \over r_i} .
\end{equation}

Hit rate is just the accuracy under a conventional classification setting and MRR is a relaxed indicator of classification performance.

\begin{table}
\centering

\scalebox{0.65}{
\begin{tabular}{ccccccc}

\hline
\multirow{2}{*}{\textbf{Approach}} & \multicolumn{2}{c}{\textbf{train}} & \multicolumn{2}{c}{\textbf{val}} & \multicolumn{2}{c}{\textbf{test}} \\
 & HIT@1 & MRR & HIT@1 & MRR & HIT@1 & MRR\\
\hline
CLIP-raw & - & - & - & - & 60.48 & 73.88 \\ 
CLIP-aug & 78.00 & 85.60 & 79.52 & 87.21 & \textbf{62.47} & \textbf{74.92} \\ 
Average fuser & 71.75 & 81.19 & 71.04 & 81.60 & 57.02 & 69.86 \\ 
MLP fuser & 78.37 &86.16 & 17.21 & 39.16 & 17.57 & 39.00\\
Transformer fuser & \textbf{89.75} & \textbf{91.63} & \textbf{81.57} & \textbf{88.80} & 57.17 & 71.64\\
\hline
\end{tabular}}
\caption{Performance of different approaches evaluated by hit rate and mean riciprocal rank (MRR).}
\label{table:results}
\end{table}

We report our results along with two baselines as shown in Table \ref{table:results}. The \texttt{CLIP-raw} baseline is provided by the organizers using zero-shot CLIP and the raw context as text input \cite{raganato-etal-2023-semeval}. Our \texttt{CLIP-aug} baseline is produced using zero-shot CLIP and the augmented context as text input. 

A counter-intuitive yet interesting result is that our baseline beats all modality fusers on the test set. One can observe that the Transformer fuser achieves strong results on both the training and validation set but its performance drops significantly on the test set. The fact that it performs well on validation data eliminates the possibility of overfitting. 
A conceivable reason for the discrepancy between the validation and test performance lies in the data. As pointed out by \citet{zhang-etal-2023-srcb} (Table \ref{tab:stat}), roughly 73.5\% of the target words in the training set are monosemic while the test set is completely the opposite, with 72.2\% words with three or more available senses. At the Gloss Matching stage, when faced with monosemic words, the bi-encoder \cite{blevins-zettlemoyer-2020-moving} will never make a single mistake because there is only one possible sense to match. However, the task for the bi-encoder gets more difficult when there are more available senses. A part of the performance drop on the test set comes from the mistakes made by the bi-encoder when matching senses. Additionally, at the Image Retrieval stage, CLIP Retrieval, either impacted by upstream errors or not, returns a percentage of images that are not strongly associated with the prompt query. Examples of such errors are shown in Appendix \ref{sec:appendix}.

Our system aggregates available knowledge to aid disambiguation with existing off-the-shelf models, instead of proposing a novel state-of-the-art WSD model. It relies on the successful execution of a chain of algorithms. Specifically, the bi-encoder is expected to return the appropriate word sense and CLIP Retrieval is expected to return images that are exact descriptions of the augmented context. Any errors along the pipeline will accumulate and hinder the final performance even with strong modality fusers. In other words, the modality fusers can only perform well if the retrieved external knowledge, either texts or images, is of high quality. Originally, we expect the modality fusers to dynamically decide which pieces of knowledge to use and which pieces to rule out. However, the huge distribution shift between the training and test data prevents them from learning to reject inaccurate knowledge since most of the knowledge retrieved when training is correct. Thus, in order for our method to excel, more robust WSD models and image-text retrieval systems should be proposed.

\begin{table}
    \centering
    \begin{tabular}{ccc}
    \hline
    \# senses & train & test \\
    \hline
     1 & \textbf{73.5\%} & 10.8\% \\
     2 & 11.9\%  & 16.9\% \\
     $\geq 3$ & 14.5\% & \textbf{72.2\%} \\
     \hline
    \end{tabular}
    \caption{Statistics on number of senses in the training set and test set, taken from \cite{zhang-etal-2023-srcb}.}
    \label{tab:stat}
\end{table}

\section{Conclusion}
Overall, our project proposes a system that collects and aggregates knowledge from different resources to aid Visual Word Sense Disambiguation. The raw context goes through a sequence of augmentation steps, including gloss matching, prompting, image retrieval, and modality fusing. Although the performance of our system is not among the best on the test set, we are able to gain insights by identifying the distributional gap between training and test data. Nevertheless, the success of our method on both the training and validation set shows its feasibility if the compositions of words of different polysemic degrees stay relatively consistent at training time and inference time. Last but not least, our work inspires researchers in the community to propose stronger models for WSD and Vision-Language representation learning.

\bibliography{custom}

\appendix

\section{Accumulated Errors along the Pipeline}
\label{sec:appendix}

This is an example where the bi-encoder first makes a mistake at matching the correct gloss and the errors are passed on so that CLIP Retrieval retrieves irrelevant images.

\vspace{5pt}
\fbox{
\begin{minipage}{18em}
\texttt{This is a picture of eating seat, where seat refers to \textcolor{red}{show to a seat; assign a seat for}.}
\end{minipage}}
\vspace{5pt}

\begin{figure*}[b]
\centering     
\subfigure[Retrieved image 1.]{\label{fig:a}\includegraphics[width=0.24\textwidth]{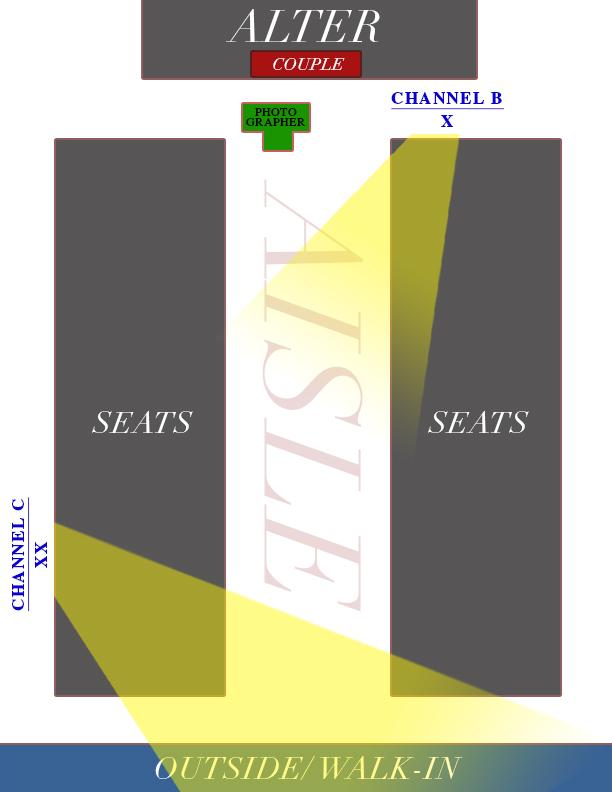}}
\subfigure[Retrieved image 2.]{\label{fig:b}\includegraphics[width=0.24\textwidth]{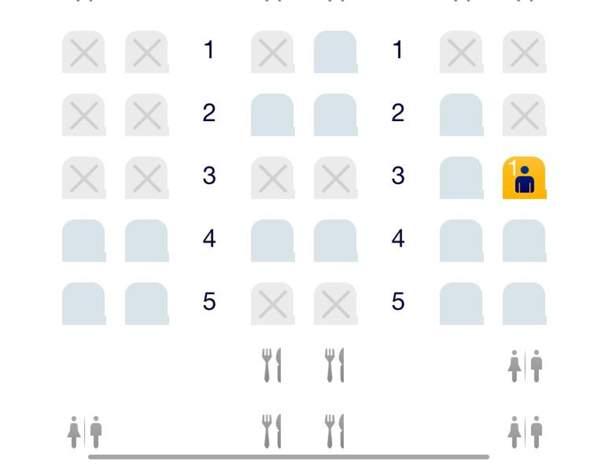}}
\subfigure[Retrieved image 3.]{\label{fig:b}\includegraphics[width=0.24\textwidth]{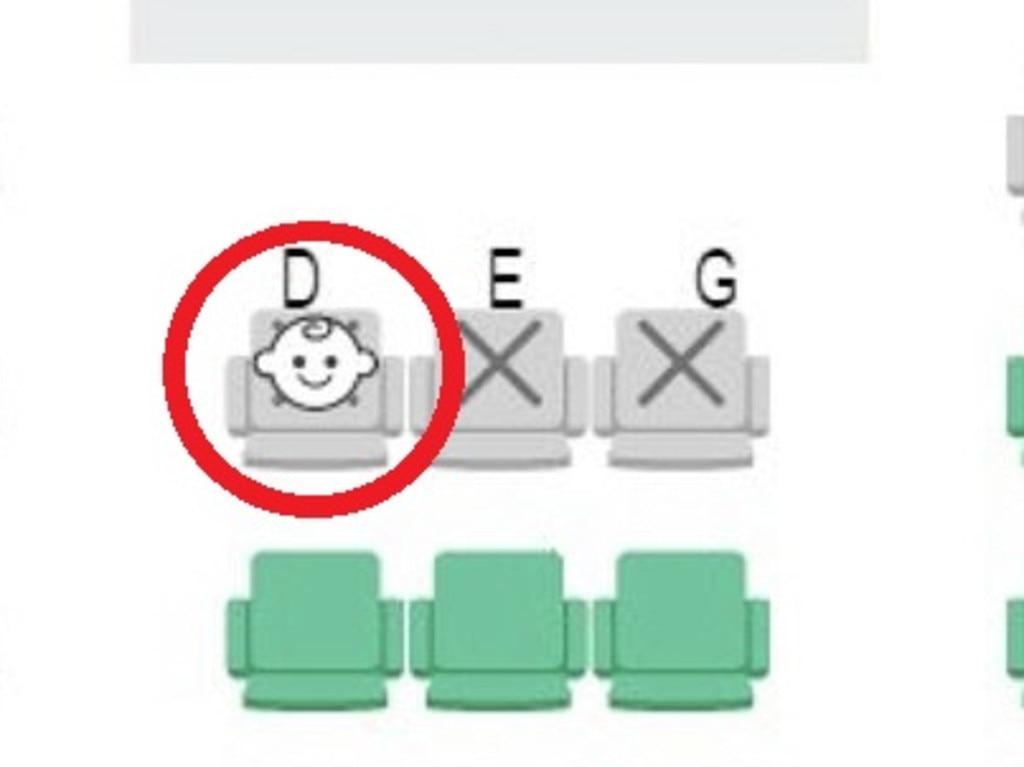}}
\subfigure[Gold image.]{\label{fig:b}\includegraphics[width=0.24\textwidth]{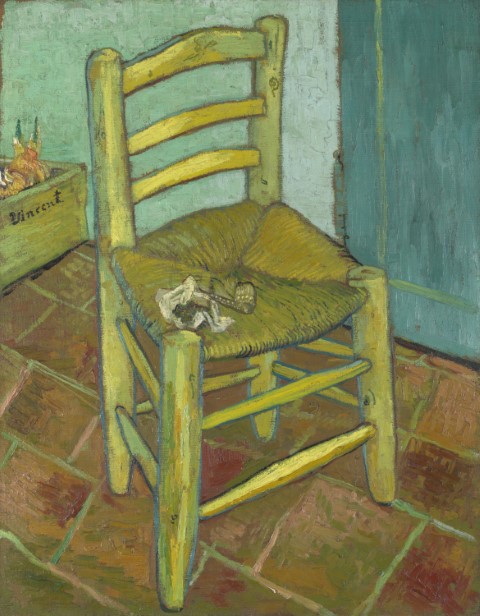}}
\caption{Retrieved images and the gold image (ground truth).}
\label{fig:fail}

\end{figure*}

The correct sense of "seat" in this context is supposed to be "furniture that is designed for sitting on". However, the bi-encoder matches it with its verb form. Naturally, the subsequent image retrieval yields reults of low relevance, as shown in Figure \ref{fig:fail}. Such cumulative errors severely hinders performance.

\end{document}